\documentclass[a4paper,conference]{IEEEtran}
\ifCLASSINFOpdf
\else
\fi
\usepackage{hyperref}
\usepackage{enumitem,kantlipsum}
\usepackage{times}
\usepackage{epsfig}
\usepackage{graphicx}
\usepackage{amsmath}
\usepackage{amssymb}
\usepackage{kotex}
\usepackage{color}
\usepackage{graphicx}
\usepackage{subcaption}
\usepackage{booktabs}
\usepackage{adjustbox}
\usepackage{kotex}
\usepackage{bbold}
\usepackage{pifont}
\newcommand{\xmark}{\text{\ding{55}}}
\usepackage{url}
\usepackage{graphicx} %Loading the package
\graphicspath{{IMAGE/}} %Setting the graphicspath

\definecolor{OliveGreen}{rgb}{0,0.6,0}

\begin{document}
\title{Feature Fusion for Online Mutual Knowledge Distillation}

% \author{\IEEEauthorblockN{Michael Shell}
% \IEEEauthorblockA{School of Electrical and\\Computer Engineering\\
% Georgia Institute of Technology\\
% Atlanta, Georgia 30332--0250\\
% Email: http://www.michaelshell.org/contact.html}
% \and
% \IEEEauthorblockN{Homer Simpson}
% \IEEEauthorblockA{Twentieth Century Fox\\
% Email: homer@thesimpsons.com}
% \and
% \IEEEauthorblockN{James Kirk\\ and Montgomery Scott}
% \IEEEauthorblockA{Starfleet Academy\\
% San Francisco, California 96678--2391\\
% Telephone: (800) 555--1212\\
% Fax: (888) 555--1212}
% \and
% \IEEEauthorblockN{James Kirk\\ and Montgomery Scott}
% \IEEEauthorblockA{Starfleet Academy\\
% San Francisco, California 96678--2391\\
% Telephone: (800) 555--1212\\


\author{\IEEEauthorblockN{Jangho Kim, Minsung Hyun, Inseop Chung and Nojun Kwak\IEEEauthorrefmark{1}} 
\IEEEauthorblockA{Email: \{kjh91, minsung.hyun, jis3613, nojunk\}@snu.ac.kr}
\IEEEauthorblockA{Seoul National University, Seoul, Korea}
\IEEEauthorblockA{\IEEEauthorrefmark{1}Corresponding Author}}


\maketitle

\begin{abstract}
We propose a learning framework named Feature Fusion Learning (FFL) that efficiently trains a powerful classifier through a fusion module which combines the feature maps generated from parallel neural networks and generates meaningful feature maps. Specifically, we train a number of parallel neural networks as sub-networks, then we combine the feature maps from each sub-network using a fusion module to create a more meaningful feature map. The fused feature map is passed into the fused classifier for overall classification. Unlike existing feature fusion methods, in our framework, an ensemble of sub-network classifiers transfers its knowledge to the fused classifier and then the fused classifier delivers its knowledge back to each sub-network, mutually teaching one another in an online-knowledge distillation manner. This mutually teaching system not only improves the performance of the fused classifier but also obtains performance gain in each sub-network. Moreover, our model is more beneficial than other alternative methods because different types of network can be used for each sub-network. We have performed a variety of experiments on multiple datasets such as CIFAR-10, CIFAR-100 and ImageNet and proved that our method is more effective than other alternative methods in terms of performances of both sub-networks and the fused classifier, and the aspect of generating meaningful feature maps. The code is available at this link\footnote{\url{https://github.com/Jangho-Kim/FFL-pytorch}}
\end{abstract}


% \fi
\IEEEpeerreviewmaketitle

\section{Introduction}

Deep neural networks have shown remarkable performance on various computer vision tasks in recent years. There have been many researches on network architecture that extracts discriminative features. In the early years, most of the works were focused on designing deeper and/or wider network to enhance the capacity of deep neural networks such as ResNet \cite{he2016deep} and wide Residual Networks \cite{zagoruyko2016wide}.

Besides developing network architecture, there have been attempts to get away from modifying the network architecture itself and to develop new training mechanism. One approach is the feature fusion method that can combine different feature maps gained from multiple sub-networks. Feature fusion methods have been used in many previous deep learning studies. In deep convolutional network models, different types of features are extracted from each layer \cite{goodfellow2016deep}. From this fact, researchers found that combining the features of each layer increases the performance of the model and showed the effectiveness of this method in various computer vision tasks such as detection, semantic segmentation and gesture classification \cite{hariharan2015hyper,fan2018multi,chang2018multi}.

DualNet \cite{how2017dual} coordinated two parallel sub-networks and trained them iteratively to learn complementary features, then they fused the two-stream features and passed it into the fused classifier. They showed that the ensemble of the fused classifier and the two classifiers of sub-networks outperforms an independently trained network.
Although DualNet uses complementary features and generates meaningful fused feature maps, directly combining feature maps incurs several challenges:

\begin{enumerate}[leftmargin=*]
    \item This approach only focuses on the performance of the fused classifier. The performance of the sub-networks is significantly lower than the performance of the network that is independently trained with the same architecture, which means ignoring the positive synergy between sub-networks and fused classifier. 
    This can also influences the performance of fused classifier.
    \item Because it directly combines feature maps, it is applicable only when the sub-networks have the same architecture.
\end{enumerate}

\begin{figure*}[t]
  \centering
  \includegraphics[width = 0.55\linewidth]{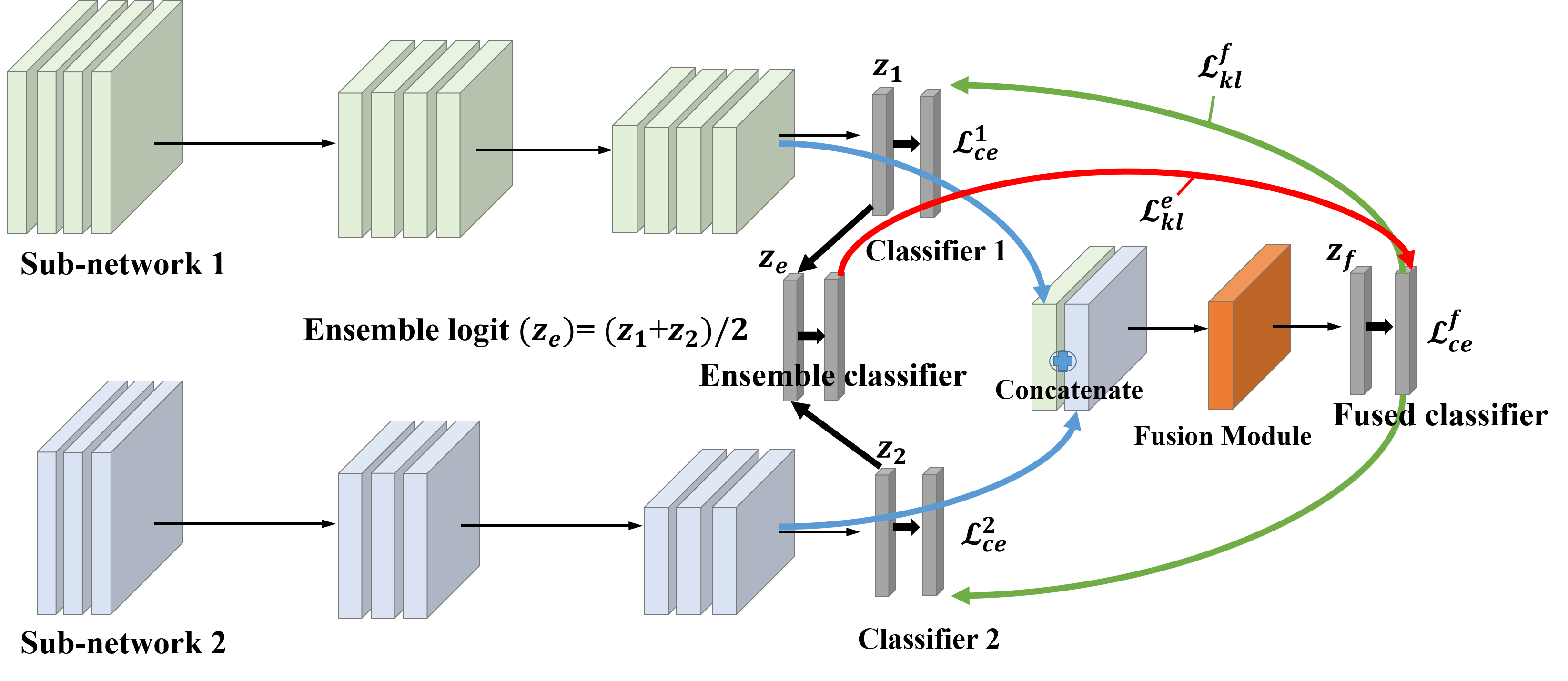}\\
  \caption{The overall process of our method is called a Feature Fusion Learning (FFL). The sub-networks create an ensemble classifier for training the fusion module. Then, the ensemble classifier transfers its knowledge to the fusion module. Similarly, the fusion module transfers its knowledge back to each sub-network. This online mutual knowledge distillation helps to obtain better performance gain in the fused classifier as well as the sub-networks. More details are explained in Sec. \ref{overallmethod}.}
    \label{fig:FFL_overall}
\end{figure*}

In this work, we propose a solution for efficiently fusing the features of sub-networks, called Feature fusion learning (FFL). Contrary to the existing feature fusion methods, we utilize the knowledge distillation \cite{hinton2015distilling} and ensemble method for making a good synergy between sub-networks and fused classifiers. Also we use the fusion module to make FFL flexible to the type of sub-network architecture. The sub-networks and the fusion module are learned by mutually teaching each other via knowledge distillation, by having a good influence on each other. This eventually creates a loop between the sub-networks and the fusion module. When the training is completed, the performances of the sub-networks as well as the fusion module are greatly improved due to the online mutual knowledge distillation between the sub-networks and the fusion module. The overall process of our method is described in Figure \ref{fig:FFL_overall}.
The contributions of this paper can be summarized as the followings:
\begin{itemize}
\item FFL can improve not only the performance of the fused classifier but also those of the sub-networks.
%classifier\nj{s} but also \nj{that} of the fused classifier.
\item Different types of sub-networks can be used in FFL, which is not the case for the existing methods. 
\item FFL can generate meaningful feature maps applicable to other tasks with online mutual knowledge distillation. 
\end{itemize}

\section{Related Work}
\subsection{Knowledge Distillation method}
Knowledge Distillation (KD) \cite{hinton2015distilling} starts with training a powerful teacher model followed by encouraging a student model to mimic the teacher model’s softened distribution. Besides probability distribution, some other researches have tried to distill various features to the student \cite{zagoruyko2016paying,kim2018paraphrasing,romero2014fitnets,yim2017gift,furlanello2018born,kim2019qkd}. In terms of training a small student network for the model efficiency, KD is also considered as an one of model compression methods such as pruning and quantization \cite{han2015deep,kim2020position}.

Deep Mutual Learning (DML) \cite{zhang2018deep} suggested a method which trains student networks to exchange information mutually through the Kullback–Leibler divergence (KLD) loss and could achieve better performance than the original network. In this framework, each student network plays the role of a teacher network to the other student networks. One advantage of this method is that any kind of different network architectures can be flexibly applied. However, this method may only provide limited information to the target because it does not utilize the rich information from the teacher model for distillation.

The On-the-fly Native Ensemble (ONE) \cite{lan2018knowledge} is one of the online distillation methods that trains only a single multi-branch network while concurrently building a strong teacher model with gating of the branch logits to enhance the learning of a student network. This method distills the knowledge of the teacher network to the student network in one-way. It uses a gating module located on the shared layer, thus it is applicable only when the branches have the same architecture. Also, this type of logit-based distillation method can not make good use of feature maps which are useful in many vision tasks. 

In our FFL, knowledge distillation is also used for enhancing the performance of sub-networks as well as the fusion module generating meaningful feature maps. 

\subsection{Feature fusion method}

The researches in \cite{lin2015bilinear,how2017dual} applied the feature fusion in dual learning. In the bilinear CNN \cite{lin2015bilinear}, outputs from two different networks are fused and mapped into a bilinear vector. DualNet \cite{how2017dual} trains two parallel networks with the same structure and uses the `SUM' operation to combine the features of those networks so as to build a fused classifier. In addition, it applies iterative training, which alternately updates the weight of the sub-networks to learn the complementary features.

\begin{figure}[t]
  \centering
  \includegraphics[width = 0.88\linewidth]{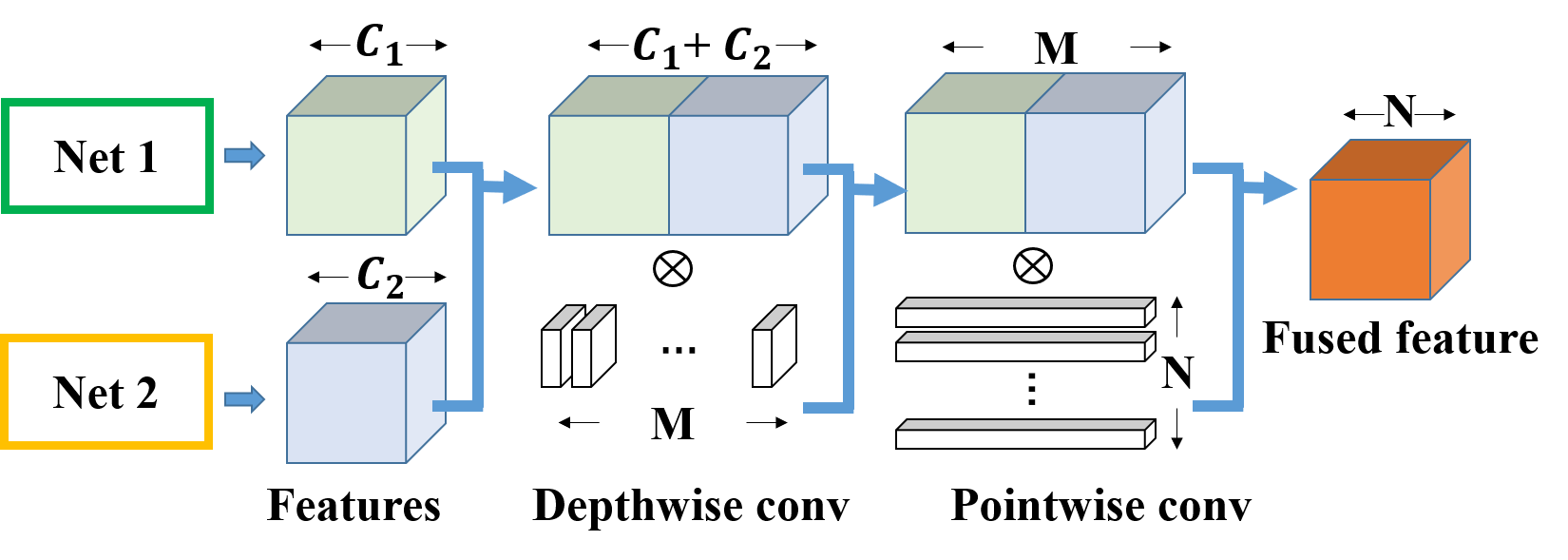}\\
  \caption{The architecture of a fusion module. The depthwise convolution is operated on concatenated feature maps of sub-networks with $M$ filters. Then, the pointwise convolution is operated with $N$ filters. }
    \label{fig:fusion}
\end{figure}

Our FFL has
three distinct points compared to DualNet. First, DualNet
is designed to work only for the same architectures of subnetworks,
whereas FFL is applicable to any network architectures.
Second, we intended the trainable fusion module to be more effective than simple feature fusion methods (See Figure \ref{fig:fusion}). 
Finally, the main difference is that DualNet is only focused on improving the performance of the fused classifier, while FFL focuses on improving
the performances of both the fused classifier and the sub-networks, using mutual knowledge distillation.

%------------------------------------------------------------------------
\section{Proposed Method}
\label{overallmethod}
In this section, we describe how to effectively fuse the features of sub-networks. The proposed method is called Feature Fusion Learning (FFL). Unlike the existing fusion methods, FFL is a learning method that takes care of not only the performance of the fused classifier but also the performance of the sub-networks. In the overall process, the features of a parallel sub-networks are fused through a fusion module, and then the final classification result is obtained through a fused classifier. During training, an ensemble of sub-networks distills its knowledge to the fused classifier, and the fused classifier distills its knowledge to each sub-network mutually.

\subsection{Fusion Module}

Different from DualNet \cite{how2017dual}, our method does not make use of the simple sum or average operation when fusing features. %with simple sum or average methods used. 
Instead, we concatenate the features of the sub-networks and then perform the convolution operation through the fusion module. To reduce the number of parameters, we use a simple depthwise convolution and an $1 \times 1$ convolution called pointwise convolution that has been used in MobileNet \cite{howard2017mobilenets}. We use the feature map of the last layer for fusion because it is specific to the task and has sufficient expressive power of the network. Let $ C_1 $ and $ C_2 $ are the numbers of channels of the feature map in the last layer of network 1 and 2, respectively, then the number of channels from the concatenated feature map, $M$, will be $ C_1 + C_2 $. The number of output channels from the fusion module, $N$, can be manipulated as needed. As shown in Figure \ref{fig:fusion}, we firstly perform a $3 \times 3$ depthwise convolution which applies a single filter per each input channel and then apply a pointwise convolution to create a linear combination of the slices of the feature map in order to combine them well.

In DualNet, there is a problem that the number of output channels of the sub-networks must be the same because the feature maps are simply averaged and added element-wise. On the other hand, in our fusion module, since the feature maps of the sub-network are concatenated, FFL can use different networks having different output channels as its sub-networks. If the resolutions of the final feature maps are different between the sub-networks, a simple convolution operation can make the spatial resolutions identical %to a small size of the sub-network resolution 
through the module which is similar to the regressor in the FitNets \cite{romero2014fitnets}.  

\subsection{Feature Fusion Learning}
\label{method}
In terms of sub-network architectures, ONE \cite{lan2018knowledge} is not flexible in that it can not be applied to sub-networks with different architectures because it creates a teacher by gating logits based on a shared feature map. Similarly, DualNet \cite{how2017dual} should also be applied to the same sub-network architecture because it simply combines features through the channel-wise sum. To overcome this problem, we designed two types of FFL %has two cases 
depending on the architectures of sub-networks in the training process:
\begin{itemize}
\item Case 1: If sub-networks have the same architecture, the low-level layers of the sub-networks are shared and the high-level layers are separated into multiple branches similar to ONE  \cite{lan2018knowledge}. 
\item Case 2: If sub-networks have different architectures, sub-networks are trained independently since sub-networks can not share the layers.
\end{itemize}

In this work, we handle the multi-class classification task. Assuming that there are $m$ classes, the logit fowarded by the $k$-th network is defined as $\mathbf{z}_k=\{z_k^1,z_k^2,...,z_k^m\}$. In the training process, we use softened probability for the model generalization. Given $\mathbf{z}_k$, the softened probability is defined as 
\begin{equation} 
\sigma_i(\mathbf{z}_k;T) = \frac{ e^{z^i_k/T}}{\sum_{j}^me^{z^j_k/T}}
\end{equation} 
When $T = 1$, it is the same as the original softmax. If the one-hot ground-truth is given as $\mathbf{y} = \{y^1,y^2,..,y^m\}$, cross-entropy loss of $k$-th network is defined as 
\begin{equation} 
\mathcal{L}_{ce}^k = -\sum_{i=1}^{m}y^{(i)}\log(\sigma_i(\mathbf{z}_k;1))
\end{equation} 

The overall process is shown in Figure \ref{fig:FFL_overall}. For illustration, we have chosen a scenario that uses different sub-network architectures (case 2). To make the most of given knowledge, we make a strong ensemble classifier. Sub-networks create the ensemble classifier through an ensemble of logits to train the fusion module. Assuming that there are $n$ sub-networks, then the ensemble of logits is computed as follows:
\begin{equation} 
\mathbf{z}_e=\frac{1}{n}\sum_{k=1}^{n}\mathbf{z}_k
\end{equation} 

To train the fusion module, the ensemble classifier distills its knowledge to the fused classifier. This is called \textit{ensemble knowledge distillation} (EKD). The EKD loss is defined as the KL-divergence between the softened distribution of the ensemble classifier and the softened distribution of the fused classifier. If the logit in the fused classifier is denoted as $ \mathbf{z}_f $, the EKD loss is as follows:
\begin{equation} 
\mathcal{L}_{kl}^e = \sum_{i=1}^{m}\sigma_i(\mathbf{z}_e;T)\log(\frac{\sigma_i(\mathbf{z}_e;T)}{\sigma_i(\mathbf{z}_f;T)})
\end{equation} 
This EKD loss can help to train a fusion module to generating meaningful feature maps compared to other methods. Feature maps from the last layer of sub-networks are concatenated and put into the fusion module. To train each sub-network, the fused classifier in the fusion module distills its knowledge to each sub-network. This is called \textit{fusion knowledge distillation} (FKD). The FKD loss for distilling the softened distribution of the fused classifier into each sub-network is defined as follows:  

\begin{equation} 
\mathcal{L}_{kl}^f = \sum_{k=1}^{n}\sum_{i=1}^{m}\sigma_i(\mathbf{z}_f;T)\log(\frac{\sigma_i(\mathbf{z}_f;T)}{\sigma_i(\mathbf{z}_k;T)})
\end{equation} 

In addition to the distillation loss, each sub-network and the fused classifier learns the true label through cross entropy and the total loss becomes 
\begin{equation} 
\mathcal{L}_{total} = \sum_{k=1}^{n}\mathcal{L}_{ce}^k+\mathcal{L}_{ce}^f+ T^2\times(\mathcal{L}_{kl}^e + \mathcal{L}_{kl}^f)
\end{equation} 

In our FFL, each sub-network and the fused classifier learns through ground-truth with cross-entropy loss. At the same time, the ensemble classifier distills its knowledge to the fused classifier with $\mathcal{L}_{kl}^e$ and in return, the fused classifier distills its knowledge to each sub-network. Through such \textit{mutual knowledge distillation} (MKD), the fusion module generates meaningful features for classification. Since the scale of the gradient produced by the softend distribution is $1/T^2$, we multiply $T^2$ according to the recommendations of \cite{hinton2015distilling}. Sub-networks and the fusion module in FFL are trained simultaneously.

\section{Experiments}

To verify our method, we compare FFL with various other methods on image classification datasets. We evaluate our method on several benchmark datasets which are CIFAR-10, CIFAR-100 \cite{krizhevsky2009learning} and ImageNet LSVRC 2015 \cite{ILSVRC15}. In Sec. \ref{ex1}, we compare our method with DualNet \cite{how2017dual}, one of the feature fusion method which has the same purpose as our method, and show the ablation study of the proposed mutual knowledge distillation method and the fusion module. Then, in Sec. \ref{ex2}, we compared FFL and Knowledge distillation methods with respect to the effectiveness of distilling the knowledge. Then, we compare our method with the ensemble method. Finally, we conduct qualitative analysis in the aspect of the feature map and generalization.

\noindent \textbf{Experiment setting:} In most experiments, we set the number of sub-networks to two, and $T=3$. In case 1, we separate the last block of a backbone network from parameter sharing and the number of output channels $ N $ of the fusion module is designed to match the smaller number of channels between $ C_1 $ and $ C_2 $. In ImageNet, we set the $N$ as $C_1 + C_2$, and separate the last 2 blocks for giving more learning capacity in the same way as \cite{lan2018knowledge}. (Sec. \ref{ex1}): We reimplemented DualNet based on the original paper and experimented by setting FFL under the same conditions as DualNet.
(Case 1 of Sec. \ref{case1}): We use the same learning schedule and hyper-parameters as in ONE. (Case 2 of Sec. \ref{case1}): For fair comparison, DML and FFL use the same learning schedule as used in ONE. 

\subsection{Comparison with Feature Fusion Method}
\label{ex1}
In this section, we compare DualNet and FFL in terms of feature fusion. Each model consists of two sub-networks with the same architecture. DualNet first trains the model with the iterative training that updates the sub-networks alternately, and then goes through the joint training process which jointly updates the sub-network classifiers and the fused classifier. On the other hand, FFL simultaneously learns two sub-networks and the fused classifier during the entire learning process. All experiments were repeated $10$ times on CIFAR-10 and CIFAR-100 datasets.

\noindent \textbf{Fused Classifier:} Table \ref{table:compare_FM_fuse} represents the top-1 error rate of the fused classifier for the test set. The performance of DualNet represents the average of classifiers, which is an ensemble of the sub-networks and the fused classifier as described in the original paper. The performance of FFL is the prediction result of the fused classifier. In CIFAR-10, FFL has slightly better performance than DualNet within the error range. Overall, as the depth of the network increases, the performance gap decreases. However, for the CIFAR-100 dataset, which is a bit more difficult problem, FFL is clearly superior to DualNet. The performance difference from ResNet-56 becomes up to $2.34\%$.

\noindent \textbf{Sub-network Classifier:} Table \ref{table:compare_FM_sub} is the top-1 error rate of all the sub-network classifiers. In this case, there are two sub-network classifiers. FFL shows better performance than DualNet and the difference is larger than that of the fused classifier experiment, because DualNet is not designed to improve the performance of sub-networks. The difference of the error rate between two methods is around $2\%$ in CIFAR-10 whereas the difference increases up to $7.85\%$ in the CIFAR-100 experiment.

Experiments on Table \ref{table:compare_FM_fuse} show that our proposed fusion module fuses features more effectively than DualNet. We also found out that FFL even improved the performance of the sub-networks which DualNet is overlooking as shown in the experiments of Table \ref{table:compare_FM_sub}. Furthermore, when using the same sub-network architecture as DualNet, FFL learns efficiently in terms of memory consumption because it uses a shared network.

\begin{table}[t]

\centering
    \begin{subtable}{.5\textwidth} \centering
        \begin{adjustbox}{width=0.95\linewidth}
    		\begin{tabular}{ l | c  c | c  c}			
                \toprule
    			 & \multicolumn{2}{|c|}{CIFAR-10} & \multicolumn{2}{|c}{CIFAR-100} \\ 
    			 $(\%)$ & DualNet & FFL & DualNet & FFL \\ \midrule
               ResNet-32 & 6.21$\pm$0.20 & 5.78$\pm$0.13 &  27.49$\pm$0.31 & 25.56$\pm$0.32\\ 
               ResNet-56 & 5.67$\pm$0.12 & 5.26$\pm$0.17 & 25.87$\pm$0.29 & 23.53$\pm$0.25  \\ 
               WRN-16-2 & 5.92$\pm$0.16 & 5.97$\pm$0.13 & 25.71$\pm$0.20 & 24.74$\pm$0.31  \\ 
               WRN-40-2 & 4.94$\pm$0.10 & 4.6$\pm$0.13 & 23.22$\pm$0.25 & 21.05$\pm$0.25  \\ 
            %     ResNet-32 & 93.79$\pm$0.20 & 94.24$\pm$0.24 &  72.51$\pm$0.31 & 74.43$\pm$0.32\\ 
            %   ResNet-56 & 94.33$\pm$0.12 & 94.74$\pm$0.17 & 74.13$\pm$0.29 & 76.47$\pm$0.25  \\ 
            %   WRN-16-2 & 94.08$\pm$0.16 & 94.07$\pm$0.17 & 74.29$\pm$0.20 & 75.26$\pm$0.31  \\ 
            %   WRN-40-2 & 95.06$\pm$0.10 & 95.40$\pm$0.13 & 76.78$\pm$0.25 & 78.95$\pm$0.25  \\ 
                  \bottomrule
            \end{tabular}
        \end{adjustbox}
             \caption{Top-1 classification error rate of fused classifiers. DualNet outputs results from the average of classifiers and FFL uses fusion module for classification.}
       \label{table:compare_FM_fuse}
    \end{subtable}

    \begin{subtable}{.5\textwidth} \centering
        \begin{adjustbox}{width=0.95\linewidth}
    		\begin{tabular}{ l | c  c | c  c}			
                \toprule
    			 & \multicolumn{2}{|c|}{CIFAR-10} & \multicolumn{2}{|c}{CIFAR-100} \\ 
    			$(\%)$ & DualNet & FFL & DualNet & FFL \\ \midrule
               ResNet-32 & 8.23$\pm$0.31 & 6.06$\pm$0.15 & 34.91$\pm$1.23 & 27.06$\pm$0.34 \\
               ResNet-56 & 7.34$\pm$0.25 & 5.58$\pm$0.13 & 32.67$\pm$1.14 & 24.85$\pm$0.30  \\ 
               WRN-16-2 & 7.53$\pm$0.20 & 6.09$\pm$0.09 & 31.7$\pm$1.00 & 25.72$\pm$0.28  \\ 
               WRN-40-2 & 6.25$\pm$0.14 & 4.75$\pm$0.16 & 28.4$\pm$0.61 & 22.06$\pm$0.20  \\ 
            %     ResNet-32 & 91.77$\pm$0.31 & 93.94$\pm$0.15 & 65.09$\pm$1.23 & 72.94$\pm$0.34 \\
            %   ResNet-56 & 92.66$\pm$0.25 & 94.42$\pm$0.13 & 67.33$\pm$1.14 & 75.15$\pm$0.30  \\ 
            %   WRN-16-2 & 92.47$\pm$0.20 & 93.91$\pm$0.09 & 68.30$\pm$1.00 & 74.28$\pm$0.28  \\ 
            %   WRN-40-2 & 93.75$\pm$0.14 & 95.25$\pm$0.16 & 71.60$\pm$0.61 & 77.94$\pm$0.20  \\ 
                  \bottomrule
            \end{tabular}
        \end{adjustbox}
            \caption{Top-1 classification error rate of sub-network classifiers.}
    \label{table:compare_FM_sub}
    \end{subtable}
    \caption{Performance comparison of two feature fusion methods, FFL and DualNet, with four different network architectures. Table \ref{table:compare_FM_fuse} is the performance of the fused classifiers. Table \ref{table:compare_FM_sub} shows the performance of sub-network classifiers. }
\label{table:compare_FM}
\end{table}

\begin{table}[ht]

\begin{adjustbox}{width=0.8\linewidth}
		\begin{tabular}{c| c c c | c  c}			
                    \toprule
        	&	&	 & & \multicolumn{2}{|c}{CIFAR-100} \\ 
        case &	FM	& EKD	&  FKD & Fused & Sub-network \\ \midrule
            A&    $\Large{\checkmark}$   &$\Large{\checkmark}$ &$\Large{\checkmark}$ &  25.56$\pm$0.32 & 27.06$\pm$0.34 \\
              B&   {$\Large{\xmark}$}  &$\Large{\checkmark}$ &$\Large{\checkmark}$ & 26.1$\pm$0.36 & 27.46$\pm$0.31  \\ 
               C&  $\Large{\checkmark}$  & {$\Large\xmark$}&$\Large{\checkmark}$ & 27.03$\pm$0.31 & 28.36$\pm$0.44  \\ 
                D& $\Large{\checkmark}$  & {$\Large\xmark$}&{$\Large\xmark$} & 27.29$\pm$0.24 & 31.04$\pm$0.31  \\ 			 
                %     FFL & 94.24$\pm$0.24 & 93.94$\pm$0.15 & 74.43$\pm$0.32 & 72.94$\pm$0.34 \\
                %   FFL$-$FM & 94.13$\pm$0.15 & 93.97$\pm$0.13 & 73.90$\pm$0.36 & 72.54$\pm$0.31  \\ 
                %   FFL$-$MKD & 94.17$\pm$0.21 & 93.40$\pm$0.21 & 72.71$\pm$0.24 & 68.96$\pm$0.31  \\ 
                      \bottomrule
        \end{tabular}
\end{adjustbox}
\centering
\caption{Ablation study of FFL. All models were trained on ResNet-32 and we evaluated the performance of each experiments with top-1 error rate on the CIFAR-100 dataset. We compared our proposed method (case A) to the cases without fusion module (case B), logit ensemble KD (case C) and fusion KD (case D).}
\label{table:FFL_ablation}
\end{table}

\noindent \textbf{Ablation Study:}
In FFL, we have taken a step forward from previous researches by introducing the fusion module (FM) and the mutual knowledge distillation (MKD) which is composed of the ensemble KD (EKD) and the fusion KD (FKD). We are going to show the efficacy of our proposed methodology through an ablation study in this part. Experiments were repeated 10 times on the CIFAR-100 dataset with two sub-networks based on ResNet-32 architecture. The numbers in Table \ref{table:FFL_ablation} represent the top-1 test error rate.

In the table, case A corresponds to our full FFL model, while case B %, \nj{the} second row of Table \ref{table:FFL_ablation}, 
is where the features are averaged like in DualNet instead of using our fusion module (FM). As expected, the error rates of the fused classifier and the sub-network classifier increase around $0.5\%$ and $0.4\%$ respectively. This is because FFL can learn by the point-wise weighted gradient that flows in the FM, acting as a gate for each feature in the concatenated feature map. Next two rows, case C and D are the cases where we remove the effect of EKD and the FKD sequentially. Without EKD (case C), the error rates of the fused and the sub-network classifiers increase by around $1.5\%$ and $1.3\%$ respectively, and EKD seems to have more influence on the fused one. When we additionally got rid of FKD (case D), the performance of the sub-network classifier shows a sharp decline compared to that of the fused classifier. This can be interpreted that the FKD has a significant impact on the performance of the sub-networks.

\subsection{Comparison with Knowledge Distillation}
In this section, to verify the effectiveness of MKD, we compared FFL with other state-of-the-art KD methods such as ONE and DML. Note that, these methods can not generate meaningful feature maps because there is no fused feature maps used in feature fusion method. Therefore, we only consider knowledge distillation perspective for comparison.
\label{ex2}

\vspace{0.5mm}
 \textbf{Case 1 : Same architecture}
\label{case1}

\noindent Since ONE \cite{lan2018knowledge} can not be applied to different sub-networks, we consider case 1 which uses sub-networks having the same architecture. ONE in the Table \ref{table:branch},\ref{table:imagenet} shows the average performance of the two branches, and ONE-E is the performance of the gated ensemble teacher. Similarly, FFL-S represents the average performance of sub-networks and FFL indicates the performance of the fused classifier.

\vspace{0.5mm}
\noindent \textbf{Branch Expansion:} FFL generally learns with two branches like DualNet. Since the Fusion module is a method that concatenates the feature maps, FFL can be learned by expanding branches like ONE. In this experiment, we apply three branches for FFL to show the possibility of expanding the branches. The experiments are conducted with ResNet-32 and ResNet-56 in CIFAR-100. All conditions were the same as ONE. Table \ref{table:branch} shows the results with 3 branch similar to those of 2 branch experiments. We can confirm that the feature fusion method also improves the performance even when the number of branches are increased.

\begin{table}[t]
\begin{adjustbox}{width=0.9\linewidth}
		\begin{tabular}{ l | c | c }			
                    \toprule
        			 & ResNet-32 & ResNet-56 \\  \midrule
                    ONE & 26.64 (26.94$\pm$0.21) \{26.61*\} & 24.63 (25.10$\pm$0.29)  \\
                   
                                       FFL-S & 26.3 (26.66$\pm$0.21)  & 24.51 (24.85$\pm$0.31) \\ \cline{1-3}

                   ONE-E & 24.75 (25.19$\pm$0.20) \{24.63*\} & 23.27 (23.59$\pm$0.24) \\ 
                     FFL & 24.31 (24.82$\pm$0.33)  & 23.20 (23.43$\pm$0.19) \\ 
                      \bottomrule
        \end{tabular}
\end{adjustbox}
\centering
\caption{Top-1 classification error rate with 3 branches. The numbers are from 10 runs of experiments and show the best values as in \protect\cite{srivastava2015training}. ``*'' represents reported results in \protect\cite{lan2018knowledge}.}
\label{table:branch}
\end{table}

\vspace{0.5mm}

\noindent \textbf{ImageNet DataSet:}
The experiments on ImageNet with ResNet-34 also have a similar tendency to those on CIFAR dataset. Both ONE and FFL have better performance than vanilla as shown in Table \ref{table:imagenet}. ONE and FFL-S have a quite similar performance. Regarding the fused classifiers, the feature based teacher shows better performance than the logits based teacher about 0.6\%. Also our FFL can generate the feature maps can be used in other tasks. This indicates that our method can be applied to a large scale realistic image dataset.

\begin{table}[t]
\begin{adjustbox}{width=0.7\linewidth}
		\begin{tabular}{ l| c | c  | c }			
                    \toprule
                    &Method & Top-1  & Top-5  \\ \hline  
                    &vanilla &26.69 &8.58 \\
                    &ONE &25.61$\pm$0.02 & 7.96$\pm$0.02  \\
                    ResNet-34&FFL-S & 25.58$\pm$0.06 &7.95$\pm$0.06  \\\cline{2-4}

                    &ONE-E & 24.48 &7.31  \\

                    &FFL & 23.91 &7.17  \\
                      \bottomrule
        \end{tabular}
\end{adjustbox}
\centering
\caption{Top-1 and Top-5 classification error rate on ImageNet. We report the average performance of two branch outputs with standard
deviation as in \protect\cite{lan2018knowledge}. \vspace{-3.5mm}}
\label{table:imagenet}
\end{table}

\begin{table}[t]
\begin{adjustbox}{width=1\linewidth}
		\begin{tabular}{ c c | c  c | c  c  }			
                \toprule
        			\multicolumn{2}{c|}{Net Types} & \multicolumn{2}{|c|}{DML} & \multicolumn{2}{|c}{FFL}  \\ 
        			Net 1 & Net 2 & Net 1 & Net 2 & Net 1 & Net 2  \\ \midrule
                  ResNet-32 & WRN-16-2 & 28.31$\pm$0.28 & 26.45$\pm$0.30 & 27.06$\pm$0.26& 25.93$\pm$0.30 \\
                                    ResNet-56 & WRN-40-2 & 26.75$\pm$0.21 & 23.33$\pm$0.27 & 26.23$\pm$0.30 & 23.06$\pm$0.43  \\
                     \bottomrule
        \end{tabular}
        \centering
\end{adjustbox}
\caption{Top-1 classification error rate on CIFAR-100. (Mean classification error (\%) of 10 runs).}
\label{table:MLM}
\end{table}

\begin{figure*}[t]
  \centering
  \includegraphics[width = 0.65\linewidth]{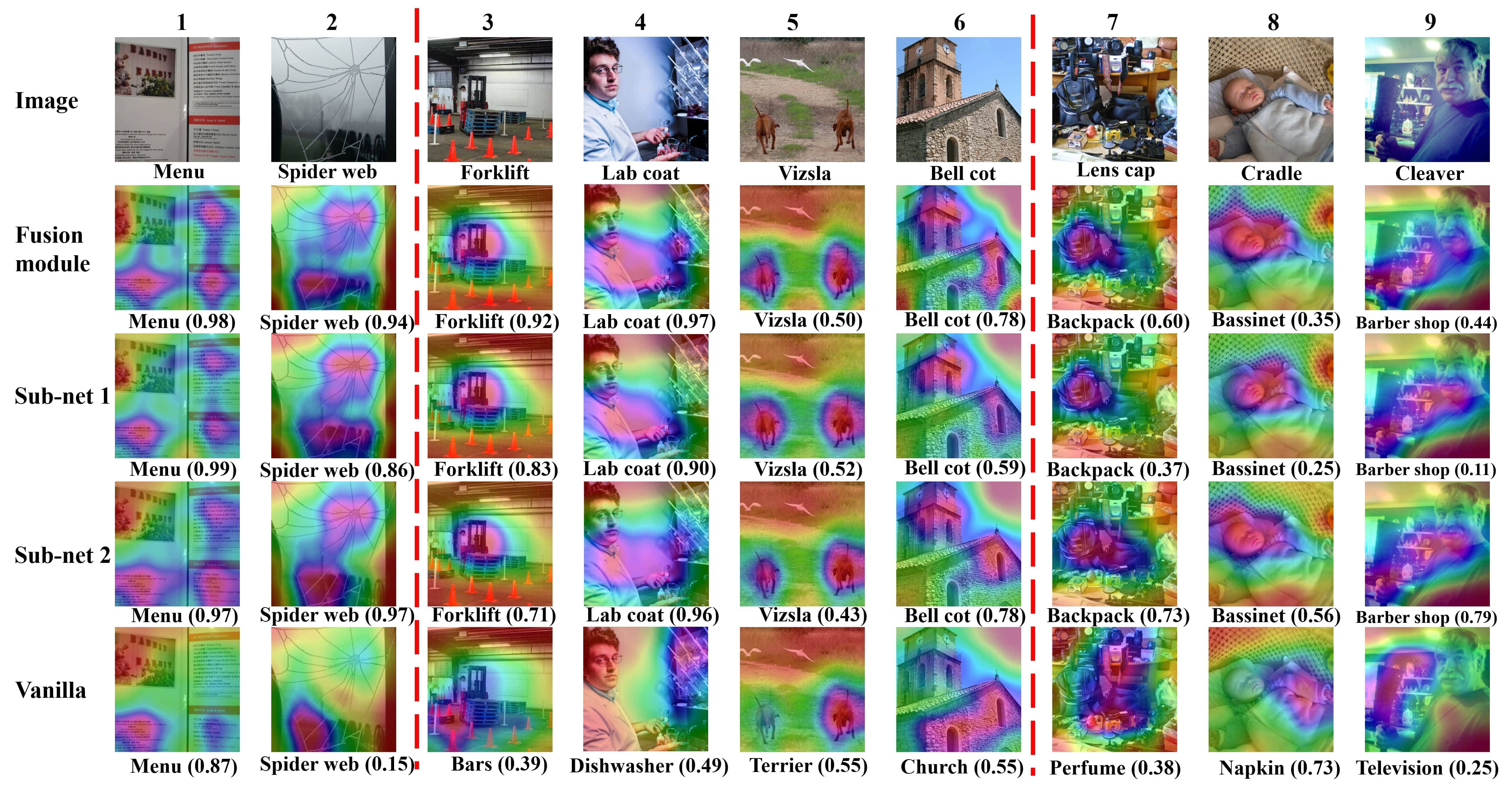}\\
  \caption{We compare the Grad-CAM \protect\cite{selvaraju2017grad} visualizations of the fusion module and the two sub-networks with the vanilla network (ResNet-34) using ImageNet dataset.}
    \label{fig:visual}
\end{figure*}

\begin{table}[t]
%\caption*{\footnotesize{Table \nj{R-A} and Table \nj{R-B} are conducted on CIFAR-100 with 3 runs. Table \nj{R-C} is conducted on ImageNet.}}
%\vspace{-4mm}
\begin{adjustbox}{width=1\linewidth}
		\begin{tabular}{ c  | c  c   c  c c c}			
                \toprule
        			 & \multicolumn{6}{c}{CIFAR-100}   \\ 
        			 & ResNet-32 & ResNet-56  & WRN-16-2 & WRN-40-2 & Pair1 & Pair2  \\ \midrule

                                    2-net \#Params & 0.94M & 1.72M & 1.40M & 4.52M  & 1.22M & 3.16M \\
                                    2-net Acc & 26.30$\pm$0.35 & 24.91$\pm$0.21 & 25.03$\pm$0.22 & 22.26$\pm$0.33 & 24.78$\pm$0.31 & 23.55$\pm$0.45 \\
                                    
                                    FFL \#Params & 0.85M & 1.54M & 1.29M  & 4.03M   & 1.19M & 3.14M\\
                                    FFL Acc & \textbf{25.45$\pm$0.28} & \textbf{24.04$\pm$0.28} & \textbf{24.70$\pm$0.33}  & \textbf{21.35$\pm$0.40} & \textbf{24.23$\pm$0.25} & \textbf{22.20$\pm$0.21} \\
                     \bottomrule
        \end{tabular}
\end{adjustbox}
\centering
\caption{Top-1 classification error rate of ensemble method (3 runs) on CIFAR-100. (Mean classification error (\%)).}
\label{table:ensemble}
\end{table}

\textbf{Case 2 : Different architectures}

\noindent In the previous experiments, sub-networks had to have the same architecture due to the architectures of the comparing methods. In case of DML, it is advantageous to be able to train sub-networks having different architectures compared to ONE. In this experiment, we compare the performance on CIFAR-100 dataset with a combination of two sub-networks having different architectures (case 2). The first combination is ResNet-34 and WRN-16-2 which has a relatively low depth and the second one is the combination of ResNet-56 and WRN-40-2 that has a deeper depth.

Table \ref{table:MLM} shows that all networks of the two combinations using FFL method is better than those of DML. FFL also obtains a stronger teacher (fused classifier) and its feature maps compared to DML. In FFL, the errors of the fused classifier for the first combination and the second combination are 24.23$\pm$0.25 and 22.20$\pm$0.21 respectively. This experiment shows that FFL method can be applied even in the case where sub-networks have different architectures. 

\vspace{0.5mm}

\noindent \textbf{Branch Expansion:} FFL can combine three different sub-nets to create meaningful feature maps and boost performances. After FFL with 3 branches (Res-32, Res-56 and WRN-16-2), each branch (same size as the vanilla) results in the error rates of 26.26$\pm$0.19, 25.21$\pm$0.38 and 25.31$\pm$0.21 respectively, which are better than vanillas by 4.7, 3.4 and 3.5\% respectively. %\ms{p}. 
With this, the fused classifier gets 22.66$\pm$0.22, which can generate fused feature maps with FM. % with fused \nj{classifier}. 

\subsection{Comparison with Ensemble method}

\vspace{0.5mm}

\noindent \textbf{CIFAR DataSet:}
We compared our method with the average ensemble network. We made the ensemble of independent networks used in (case 1) and (case 2) called as 2-net in Table \ref{table:ensemble}.
In case 1, 2-net has more parameters compared to FFL because FFL shares low-level parameters. In Pair1 (Res-32\&WRN-16-2) and Pair2 (Res-56\&WRN-40-2) of case 2, we stack more blocks for 2-net to match the parameter size with FFL. 

\vspace{0.5mm}

\noindent \textbf{ImageNet DataSet:}
We also compared to the ensemble method which uses two ResNet-34 on ImageNet. 2-net and sub-net error rates are 24.36 and 26.37$\pm$0.01, respectively. Our numbers are in Table \ref{table:imagenet}.

FFL is always better than 2-net ensemble with fewer parameters in every case and every dataset. These results indicate that a classifier using the fused feature is better than the classifier using the ensemble method of two independent networks.

\subsection{Qualitative analysis}
\label{ex4}

We aim to give insights on how our FFL method is contributing to the performance of our model by analyzing the feature map outputs. We have created heatmaps of features from four different networks which are the fusion module, the two sub-networks and an independently trained ResNet-34 network. We applied Grad-CAM \cite{selvaraju2017grad} algorithm which is a method that visualizes the important regions where the network has considered important to discover how our model is making use of the features. Figure \ref{fig:visual} shows visualizations from each network with the highest probability and the corresponding class. Columns 1-2 show cases where both the networks of our model and ResNet-34 predict the correct class. Columns 3-6 are cases where ours get the correct answer but the vanilla does not. Columns 7-9 show that the feature maps of the fusion module and the sub-networks are very similar and predict the same class even when they get wrong answer. We have observed that the networks of our model detect the correct object better than the vanilla ResNet-34. Even when both the vanilla and our three networks predict the same correct answer, ours have higher rate of confidence (First two columns of Figure \ref{fig:visual}). Also, we have discovered that the heatmaps of the sub-networks have a tendency to mimic the heatmap of the fusion module. This implies that the sub-networks are greatly influenced by the fusion module and vice versa. This is mainly due to the mutual knowledge distillation between the sub-networks and the fusion module which transfers softened probabilities that has rich information about the relative probabilities of incorrect answers. Due to the  fusion module, FFL can create meaningful feature maps that can be used in various vision tasks \cite{ren2015faster,chen2017rethinking,johnson2016perceptual} compared to other KD methods such as ONE and DML, and the ensemble method.

\section{Conclusion}
We propose a feature fusion method using online mutual knowledge distillation. Unlike existing feature fusion methods, it focuses on not only the performance of the fused classifier but also the performance of the sub-networks and can deploy sub-networks as needed. Moreover, there is no constraint on the architecture of the sub-networks. Therefore, the features of different sub-networks can be fused. The fusion module generates meaningful features. From various perspectives, we demonstrated the effectiveness of FFL in three datasets.

\bibliographystyle{IEEEtran}
\bibliography{FFL}

\clearpage
\onecolumn

\appendix
\section{Implementation details}
All experiments were conducted in the Torch framework. We used ResNet \cite{he2016deep} and wide residual networks (WRN) \cite{zagoruyko2016wide} architectures for several experiments. The fusion module consists of a single $3 \times 3$ depthwise convolution layer and a single $1 \times 1$ pointwise convolution layer. The fusion module used both batchnorm and ReLU nonlinearities for both layers.   
Similar to \textit{burn in steps} in Codistillation \cite{anil2018large},
in our method, each student network is trained somewhat independently before some epochs so as to limit the distillation at the beginning of the learning using the weight ramp-up function proposed by \cite{laine2016temporal}, which is also used in On-the-fly Native Ensemble (ONE) \cite{lan2018knowledge}.

\subsection{CIFAR Dataset}
We used the data augmentation including horizontal flips and random crops from an image padded by 4 pixels in all CIFAR experiments. The proposed fusion module was used with lower weight decay than that of sub-networks because high weight decay regularizes too much for the fusion module causing low training accuracy of a fused classifier. Detailed experimental settings are listed in Table \ref{table:exp_cifar}.

\vspace{3mm}
\noindent \textbf{Comparison with Feature Fusion Method (vs DualNet):} We reimplemented the released code of DualNet \footnote{\url{https://github.com/ustc-vim/dualnet}} for the Torch framework. Sub-networks of DualNet are trained alternately for 256 epochs first (iterative training) and then trained jointly for the rest of the epochs (joint training). The average classifier of DualNet is defined as follows:
\begin{equation} 
Z_{avg} = \lambda_1 * Z_{sub1} + \lambda_2 * Z_{sub2} + Z_{fused}
\label{average_classifier}
\end{equation} 
where $\lambda_1 = \lambda_2 = 0.3$, $Z_{sub1}, Z_{sub2}$ are the logits of sub-network classifiers and $Z_{fused}$ is the logit of the fused classifier.

\begin{table*}[t]
\centering
\caption{Experiment settings in the comparison with Feature Fusion Leaning.}
% \begin{adjustbox}{width=0.5\linewidth}
		\begin{tabular}{l|c | c}			
        \toprule
        Parameter &	vs DualNet & vs ONE,vs DML \\ \midrule
        \# of sub-network & 2  & 2\\
        Training epoch & 307  & 300\\
        Optimizer & Nesterov SGD &  Nesterov SGD\\
        Learning rate & $0.1\xrightarrow{}0.01$ & $0.1\xrightarrow{}0.01\xrightarrow{}0.001$\\
        Learning rate decay epoch & 256 & 150,225\\
        Momentum & 0.9  &0.9\\

        Weight decay & $1e\text{-}4$ &$1e\text{-}4$ \\
        Weight decay of the fusion module & $1e\text{-}5$ &$1e\text{-}5$ \\

        Weight ramp-up epoch & 80 &80 \\
        Train batch size & 128  &128\\
        Test batch size & 100  &100\\
        \bottomrule
        \end{tabular}
% \end{adjustbox}
\label{table:exp_cifar}
\end{table*}

\vspace{3mm}
\noindent \textbf{Comparison with Online ensemble Distillation (vs ONE):} We used the same schedules and hyper-parameters as used in \cite{lan2018knowledge}. We used the released code of ONE\footnote{\url{https://github.com/Lan1991Xu/ONE_NeurIPS2018}} for comparisons. For training the sub-networks, we used the SGD with Nesterov momentum with 0.9 and the weight decay was set to $10^{-4}$. The learning rate drops from 0.1 to 0.01 at 150 epoch and to 0.001 at 225 epoch. 

\vspace{3mm}
\noindent \textbf{Comparison with Mutual Learning Method (vs DML):} For fair comparisons, we used the same learning schedules and hyper-parameters as ``Comparison with Online ensemble Distillation'' for FFL and Deep mutual learning (DML) \cite{zhang2018deep}. The only differences are using the different architecture of sub-networks and that DML iteratively updates parameters.

\subsection{ImagNet Dataset}
We also used the data augmentation including horizontal flips and random $224 \times 224$ crops from an image. In the training of the sub-networks, we started with a learning rate of 0.1. The learning rate is decayed by a factor of 0.1 at every 30 epochs, as typical setting in the
ImageNet training. We stopped the training process at 90 epochs. We used the SGD with Nesterov momentum with 0.9 and the weight decay was set to $10^{-4}$, with the mini-batch size of 256. Similar to CIFAR dataset, we set the weight decay of the fusion module as $10^{-5}$. We set the weight ramp-up epoch as 20.

\section{Qualitative analysis}

\begin{figure*}[h]
  \centering
  \includegraphics[width = 1\linewidth]{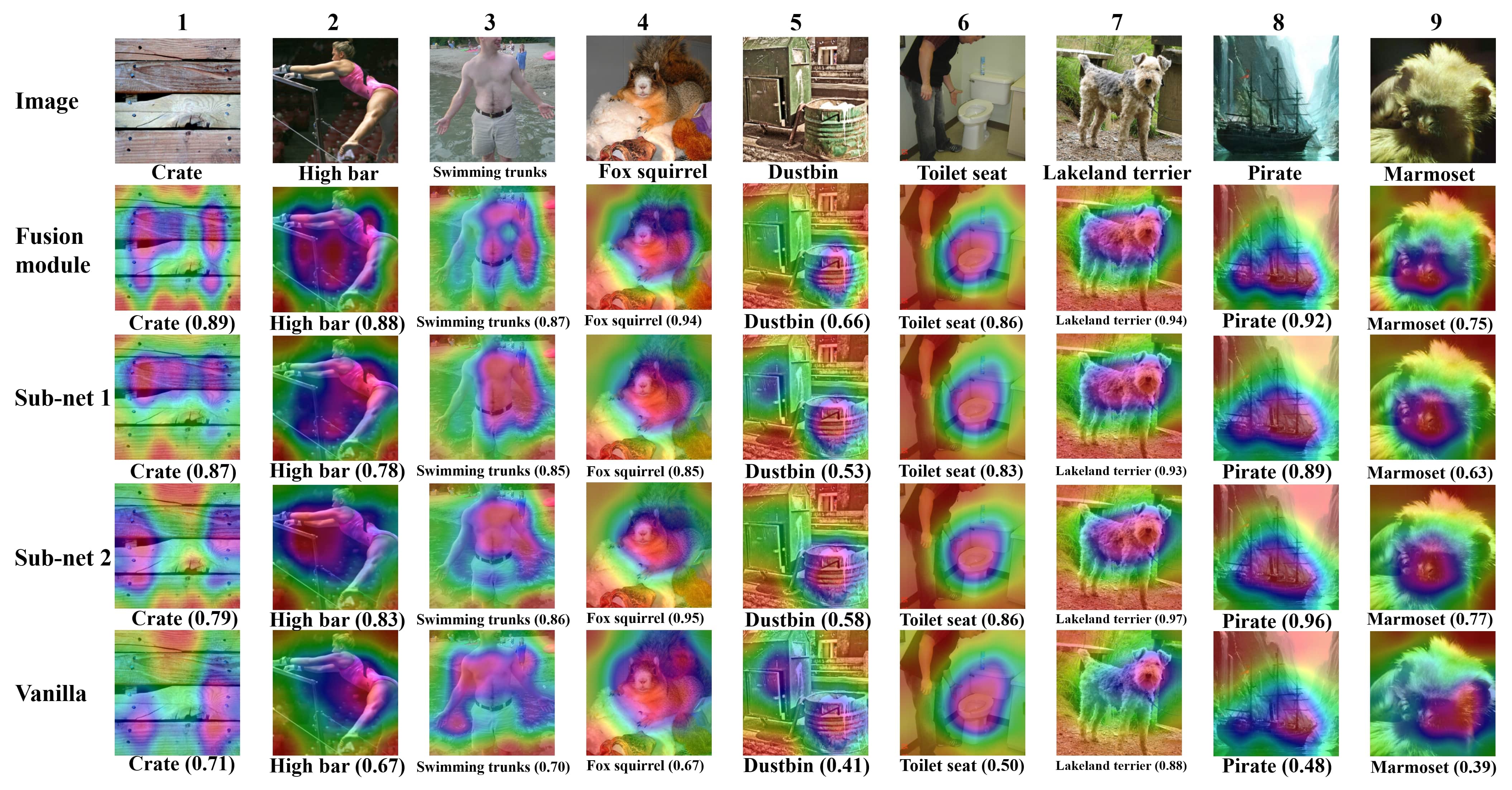}\\
  \caption{We compare the Grad-CAM \cite{selvaraju2017grad} visualizations of the fusion module and the two sub-networks with the vanilla network (ResNet-34) using ImageNet dataset (case 1).}
    \label{fig:case1_1}
\end{figure*}

\begin{figure*}[h]
  \centering
  \includegraphics[width = 1\linewidth]{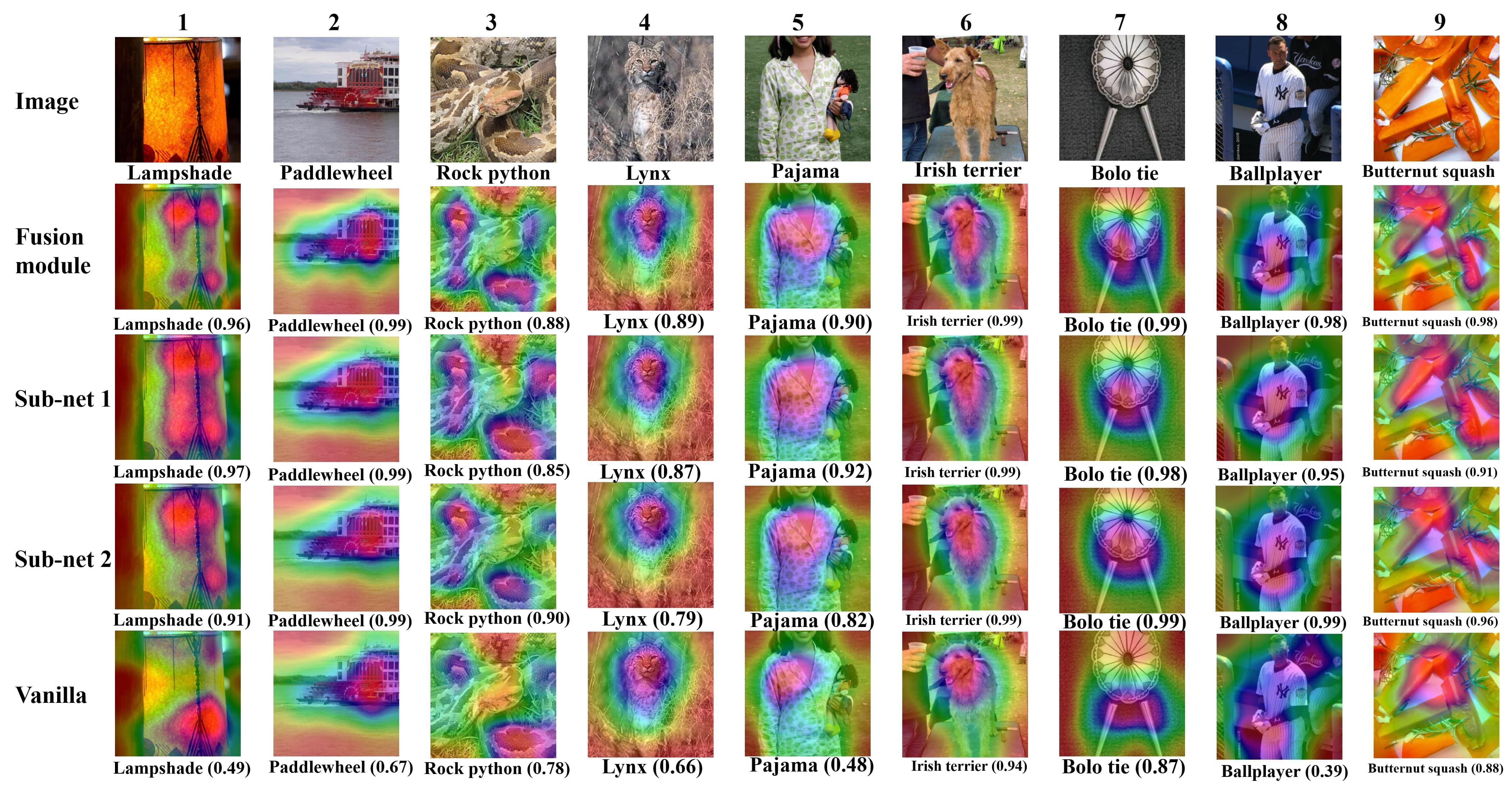}\\
  \caption{We compare the Grad-CAM \cite{selvaraju2017grad} visualizations of the fusion module and the two sub-networks with the vanilla network (ResNet-34) using ImageNet dataset (case 1).}
    \label{fig:case1_2}
\end{figure*}

\begin{figure*}[h]
  \centering
  \includegraphics[width = 1\linewidth]{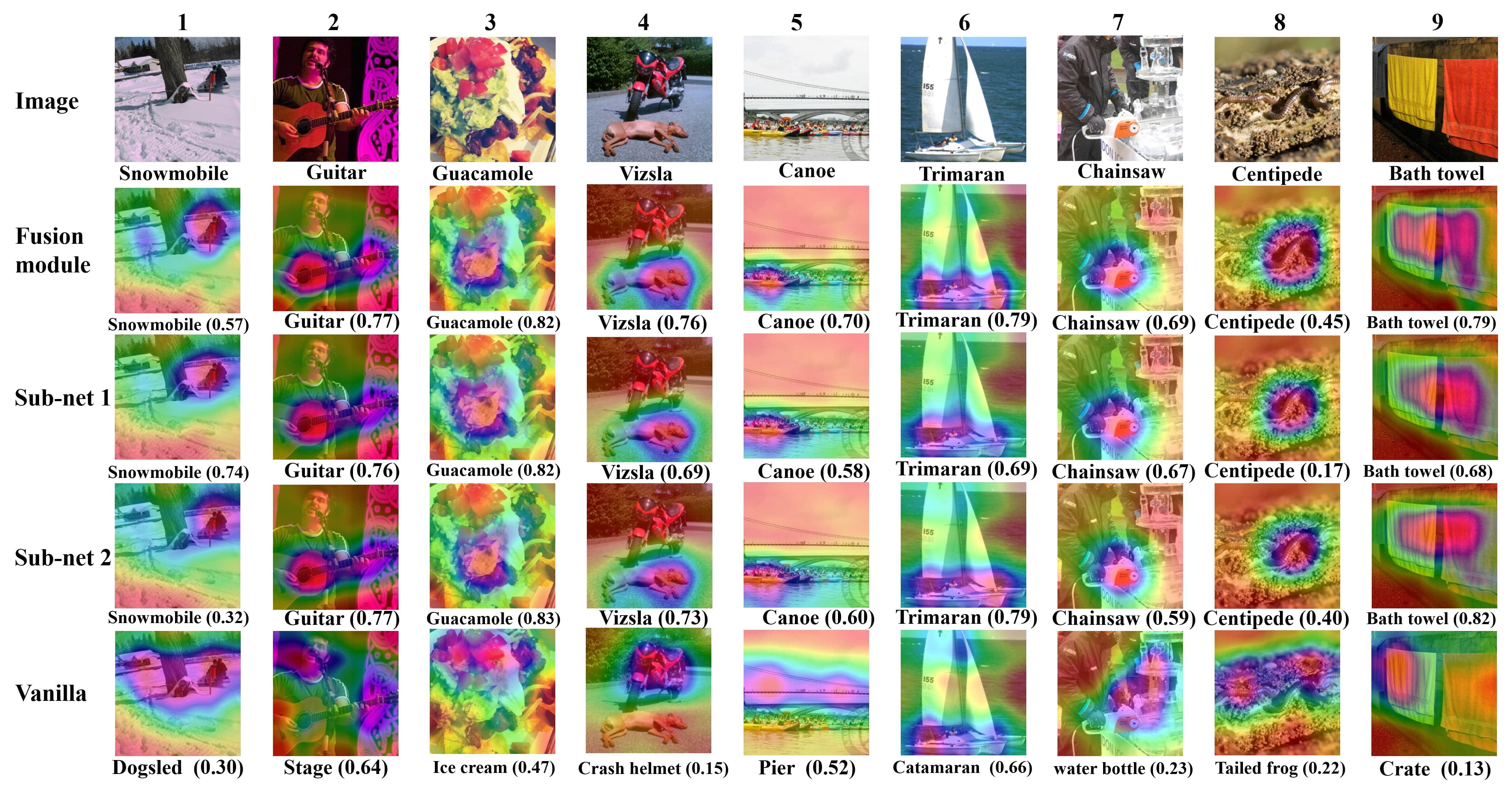}\\
  \caption{We compare the Grad-CAM \cite{selvaraju2017grad} visualizations of the fusion module and the two sub-networks with the vanilla network (ResNet-34) using ImageNet dataset (case 2)}
    \label{fig:case2_1}
\end{figure*}

\begin{figure*}[h]
  \centering
  \includegraphics[width = 1\linewidth]{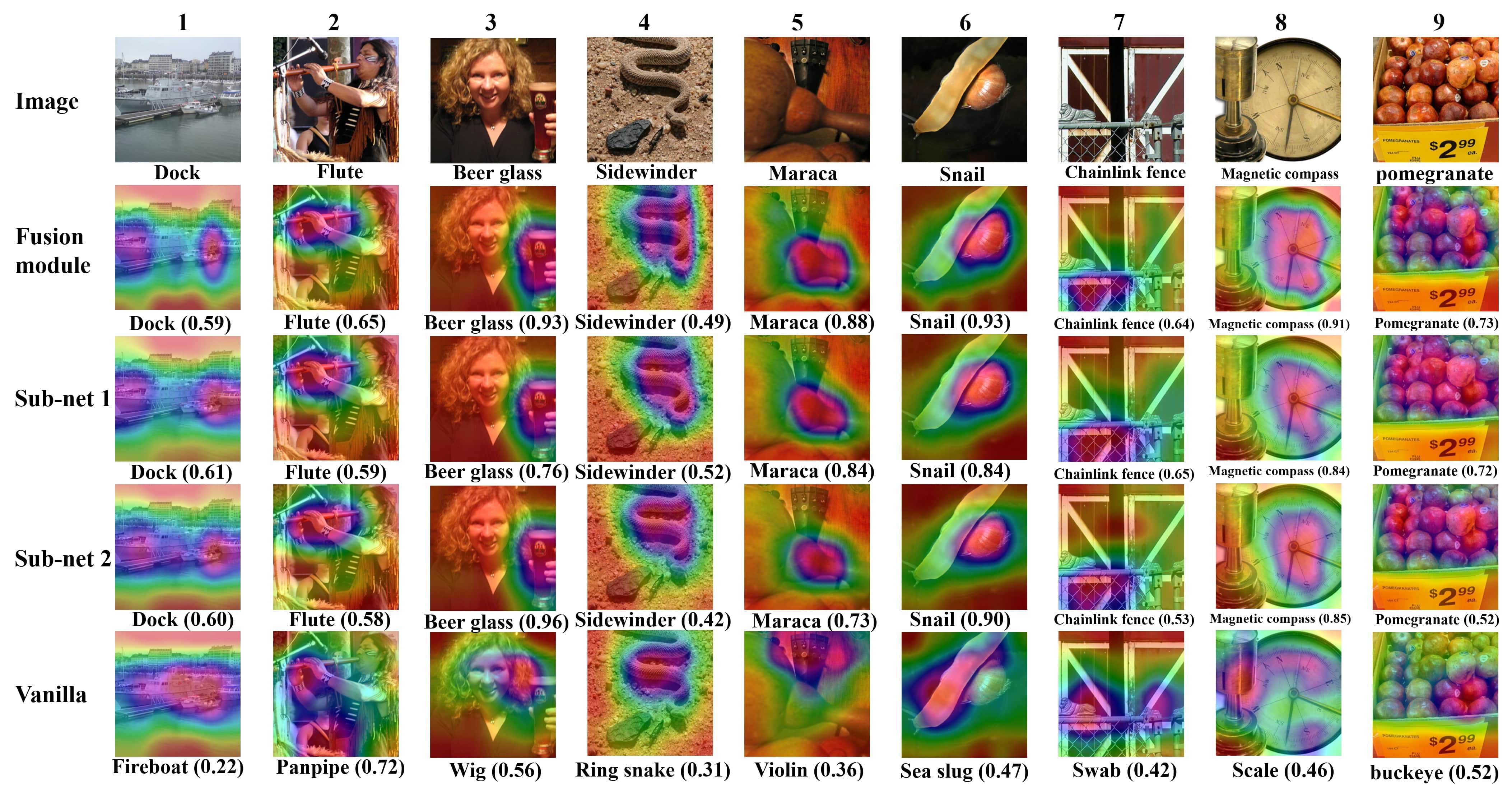}\\
  \caption{We compare the Grad-CAM \cite{selvaraju2017grad} visualizations of the fusion module and the two sub-networks with the vanilla network (ResNet-34) using ImageNet dataset (case 2)}
    \label{fig:case2_2}
\end{figure*}

\begin{figure*}[h]
  \centering
  \includegraphics[width = 1\linewidth]{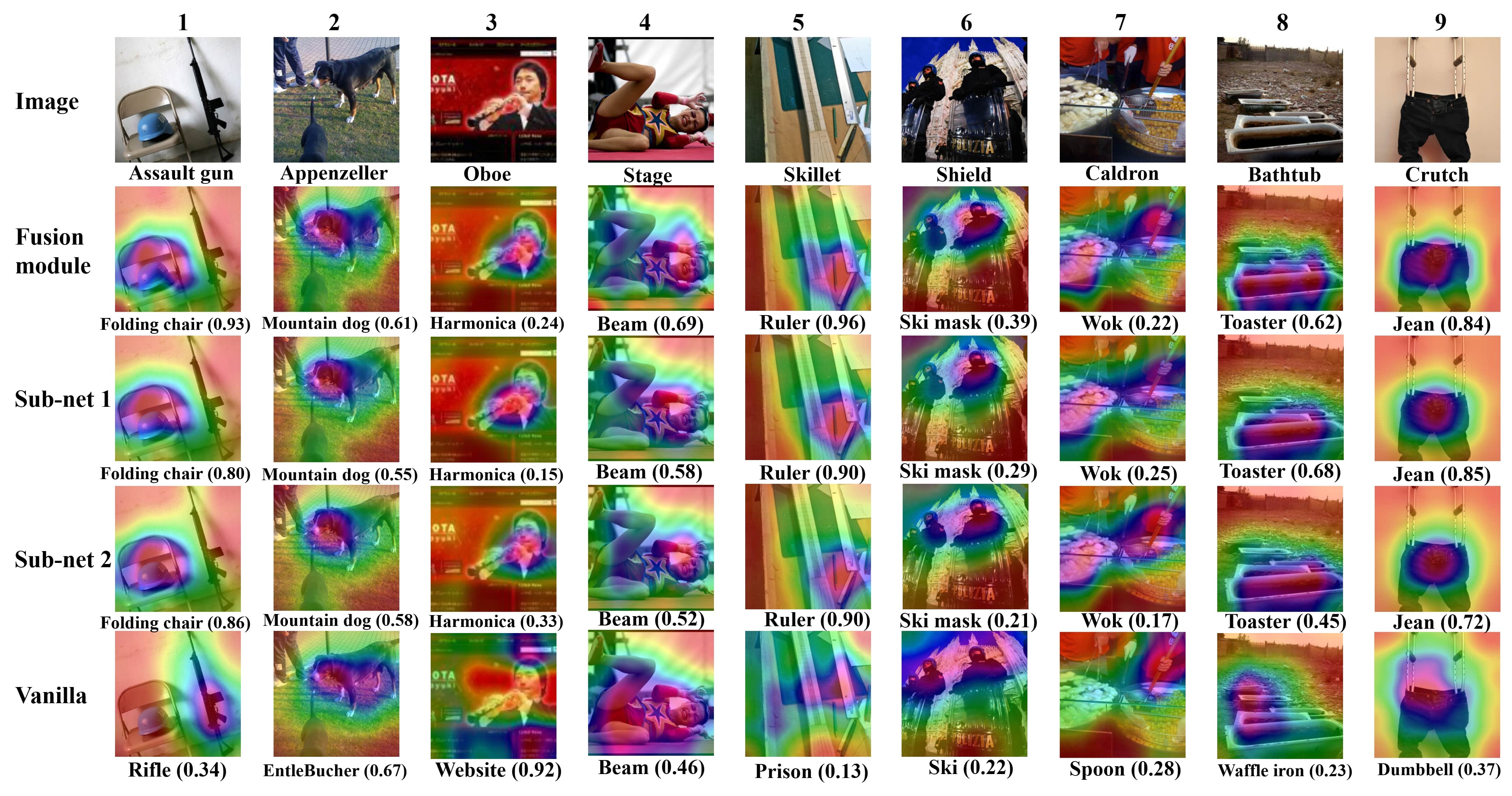}\\
  \caption{We compare the Grad-CAM \cite{selvaraju2017grad} visualizations of the fusion module and the two sub-networks with the vanilla network (ResNet-34) using ImageNet dataset (case 3)}
    \label{fig:case3_1}
\end{figure*}

\begin{figure*}[h]
  \centering
  \includegraphics[width = 1\linewidth]{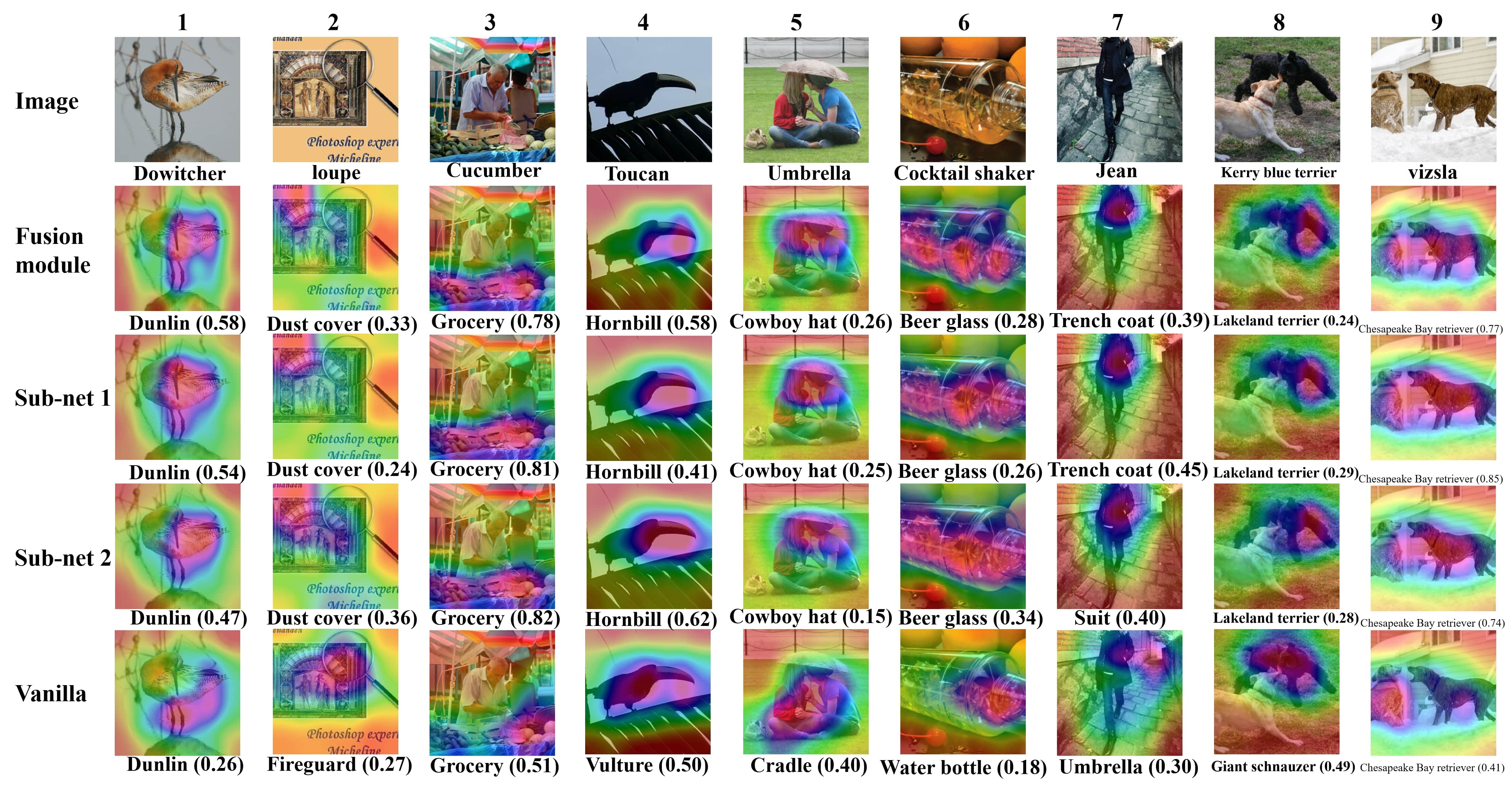}\\
  \caption{We compare the Grad-CAM \cite{selvaraju2017grad} visualizations of the fusion module and the two sub-networks with the vanilla network (ResNet-34) using ImageNet dataset (case 3)}
    \label{fig:case3_2}
\end{figure*}

We attached more results of Grad CAM visualizations on ImageNet dataset using the fusion module, sub networks and the vanilla network (ResNet-34). Figure \ref{fig:case1_1},\ref{fig:case1_2} show that all four networks have a correct answer (case 1) and the Figure \ref{fig:case2_1}, \ref{fig:case2_2} show that the fusion module and sub-networks have a correct answer but vanilla has a wrong answer (case 2). Finally, Figure \ref{fig:case3_1}, \ref{fig:case3_2} show all networks have a wrong answer (case 3).

\end{document}